\documentclass[11pt]{article}

\usepackage[preprint]{acl}

\usepackage{times}
\usepackage{latexsym}

\usepackage[T1]{fontenc}

\usepackage[utf8]{inputenc}

\usepackage{microtype}

\usepackage{inconsolata}

\usepackage{graphicx}

\usepackage{enumitem}
\usepackage{tabularx}
\usepackage{multirow}
\usepackage{booktabs}
\usepackage{tikz}
\usepackage{hyperref}
\usepackage[table]{xcolor}
\usepackage{arydshln}
\usepackage{amsmath}
\usepackage{amsfonts}
\usepackage{amssymb} 
\usepackage{CJKutf8}

\title{TraceSIR: A Multi-Agent Framework for Structured Analysis and Reporting of Agentic Execution Traces}

\author{
Shu-Xun Yang$^{1,2}$\thanks{Equal contribution. Work done at Zhipu AI.}  \quad
Cunxiang Wang$^{2,3 *}$\thanks{The corresponding author} \quad
Haoke Zhang$^{2}$ \quad
Wenbo Yu$^{2}$ \quad
Lindong Wu$^{2}$ \\
\textbf{Jiayi Gui}$^{2}$ \quad
\textbf{Dayong Yang}$^{2}$ \quad
\textbf{Yukuo Cen}$^{2}$ \quad
\textbf{Zhuoer Feng}$^{2,3}$ \quad
\textbf{Bosi Wen}$^{2,3}$ \\
\textbf{Yidong Wang}$^{2}$ \quad
\textbf{Lucen Zhong}$^{2}$ \quad
\textbf{Jiamin Ren}$^{2}$ \quad
\textbf{Linfeng Zhang}$^{4}$ \quad
\textbf{Jie Tang}$^{3}$ \\
$^{1}$Beijing Institute of Technology, Beijing, China \quad
$^{2}$Zhipu AI, Beijing, China \\
$^{3}$Tsinghua University, Beijing, China \quad
$^{4}$Shanghai Jiao Tong University, Shanghai, China \\
\texttt{sheryl.xun@bit.edu.cn; wangcunxiang303@gmail.com}
}

\begin{document}
\maketitle
\begin{abstract}

Agentic systems augment large language models with external tools and iterative decision making, enabling complex tasks such as deep research, function calling, and coding. However, their long and intricate execution traces make failure diagnosis and root cause analysis extremely challenging. Manual inspection does not scale, while directly applying LLMs to raw traces is hindered by input length limits and unreliable reasoning. Focusing solely on final task outcomes further discards critical behavioral information required for accurate issue localization.
To address these issues, we propose \textbf{TraceSIR}, a multi-agent framework for structured analysis and reporting of agentic execution traces. TraceSIR coordinates three specialized agents: (1) StructureAgent, which introduces a novel abstraction format, \texttt{TraceFormat}, to compress execution traces while preserving essential behavioral information; (2) InsightAgent, which performs fine-grained diagnosis including issue localization, root cause analysis, and optimization suggestions; (3) ReportAgent, which aggregates insights across task instances and generates comprehensive analysis reports.
To evaluate TraceSIR, we construct \textbf{TraceBench}, covering three real-world agentic scenarios, and introduce \texttt{ReportEval}, an evaluation protocol for assessing the quality and usability of analysis reports aligned with industry needs. Experiments show that TraceSIR consistently produces coherent, informative, and actionable reports, significantly outperforming existing approaches across all evaluation dimensions. 
Our project and video are publicly available at https://github.com/SHU-XUN/TraceSIR.

\end{abstract}

\section{Introduction}

\begin{figure}[tb]
  \centering
  \includegraphics[width=\linewidth]{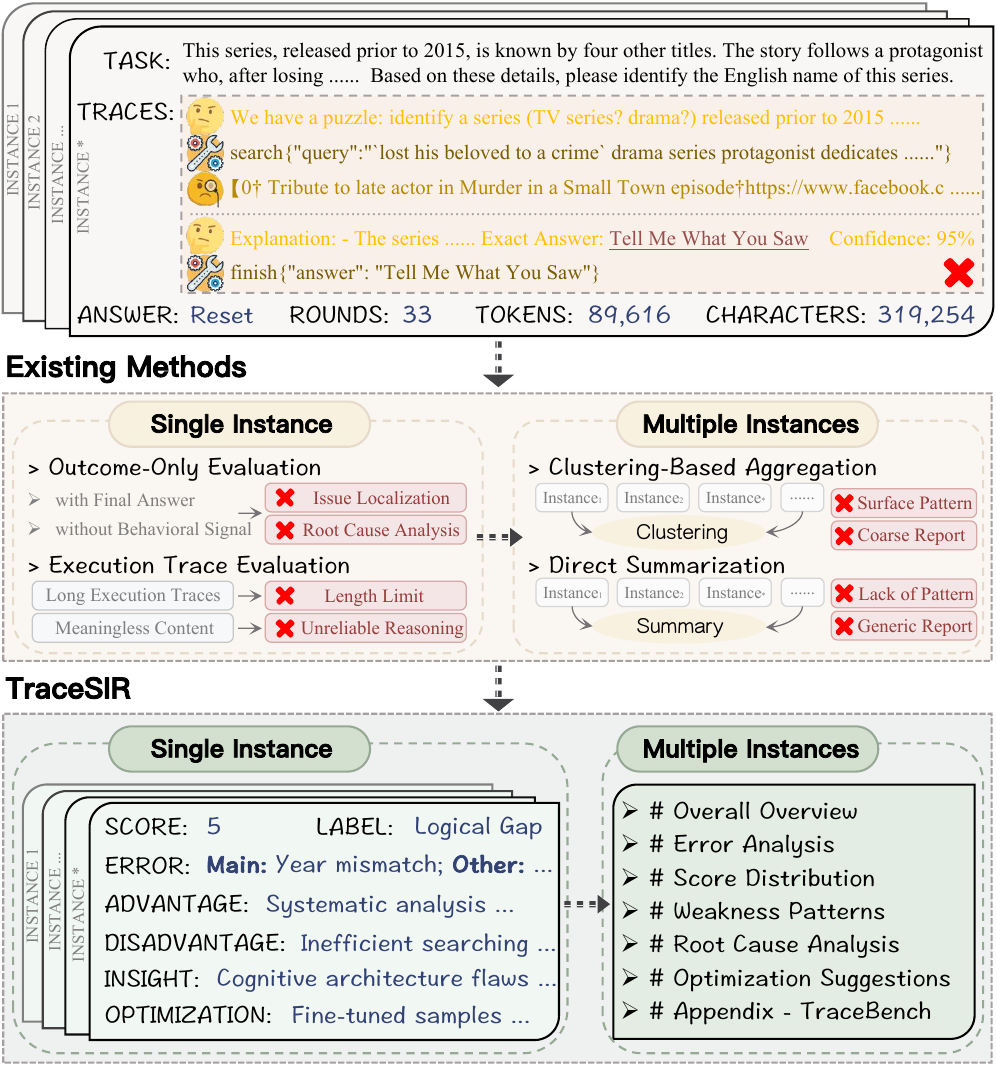}
  \caption{Comparison of existing methods and \textbf{TraceSIR} for agentic execution trace analysis and reporting.}
  \label{fig: motivation}
  \vspace{-5mm}
\end{figure}

Agentic systems, in which large language models (LLMs) are augmented with external tools and iterative decision making, have emerged as a powerful paradigm for real-world applications such as ClaudeCode and OpenClaw. As a result, improving the effectiveness of agentic systems has become an important research topic. This requires the identification of the limitations of existing agentic models and the root causes of the issues they exhibit. However, the dominant practice of manually inspecting agent outputs does not scale to the long and complex execution traces of agentic systems, where a single task may involve thousands of tool invocations and sub-agent interactions spanning large numbers of tokens, making comprehensive understanding and root cause analysis prohibitively difficult for humans \citep{trail}. 
This difficulty is further compounded by the fact that valuable issue localization and root cause analysis often require coordinated comparison across multiple task instances. Such cross-instance analysis substantially increases the analytical burden and underscores the urgent need for automated analysis and reporting frameworks.

However, automated analysis of agentic systems faces a fundamental bottleneck, as long execution traces can easily exceed the input length limits of LLMs and substantially interfere with their ability to reliably analyze agent behavior, often leading to meaningless or hallucinatory content \citep{trace,mas-fire}, as shown in Figure \ref{fig: motivation}.
One possible alternative is to focus solely on final task outcomes, but relying only on outcomes discards large amounts of valuable information embedded in the execution traces, which directly hinders accurate issue localization and prevents root cause analysis that depends on behavioral context.
Accordingly, a central challenge is how to abstract agentic execution traces in a way that preserves core information necessary for accurate issue localization and root cause analysis. 
In addition, presenting researchers with comprehensive and coherent analysis reports poses another important challenge. To date, little prior work has focused on generating and evaluating analysis reports that align with practical industry needs.

To address these challenges, we propose \textbf{TraceSIR}, a multi-agent framework for structured analysis and reporting of agentic execution traces. Specifically, TraceSIR coordinates three specialized agents that operate collaboratively, namely \textit{StructureAgent}, \textit{InsightAgent}, and \textit{ReportAgent}. \textit{StructureAgent} introduces a novel abstraction representation, referred to as \texttt{TraceFormat}, which structurally abstracts execution traces to substantially reduce redundancy while preserving core information from the original traces. Built on the structured traces, \textit{InsightAgent} performs fine grained analysis, including overall assessment, issue localization, weakness identification, root cause analysis, and optimization suggestions. Finally, \textit{ReportAgent} summarizes critical errors observed in \textit{InsightAgent} and determines whether report generation is warranted based on the number of task instances. When triggered, it conducts targeted statistical analysis and produces comprehensive analysis reports that provide researchers and engineers with systematic and detailed insights to support decisions.


To evaluate the effectiveness of TraceSIR, we collect task instances from three representative real-world agentic benchmarks, including BrowseComp (Deep Research) \citep{browsecomp-deepresearch}, Tau2Bench (Function Calling) \citep{tau2-functioncalling}, and SWE-bench (Agentic Coding) \citep{swebench-coding}, and construct a unified benchmark named \textbf{TraceBench} for systematic experimental analysis. Specifically, we select 50 failure cases from each benchmark using GLM-4.6 as the target model, and retain execution traces in the standard OpenAI message format, which serve as inputs to our TraceSIR for structured analysis and reporting of agentic execution traces. 
To reflect the practical needs of researchers and engineers in real industry settings, we further introduce a report evaluation protocol, referred to as \texttt{ReportEval}. ReportEval defines a set of gold standard principles for assessing analysis reports and adopts LLM-as-a-judge to assist human annotation, enabling end-to-end evaluation of the quality and usability of reports generated by TraceSIR.

Experiments under the \texttt{ReportEval} protocol show that the analysis reports generated by TraceSIR on TraceBench consistently outperform those produced by the strong baseline ClaudeCode. On average, TraceSIR improves overall report quality by 9.7\% under human evaluation by expert LLM Agent researchers and 7.5\% under LLM-as-a-judge evaluation, with the largest relative gain reaching 26.0\%. These results indicate that TraceSIR produces analysis reports that better align with practical industry needs and deliver more reliable and informative diagnostic insights.
More importantly, TraceSIR yields clear and robust advantages in issue localization, error interpretation, root cause analysis, and optimization suggestions, demonstrating its effectiveness in trace structuring and analysis across different underlying LLMs.


\begin{figure*}[tb]
  \centering
  \includegraphics[width=\linewidth]{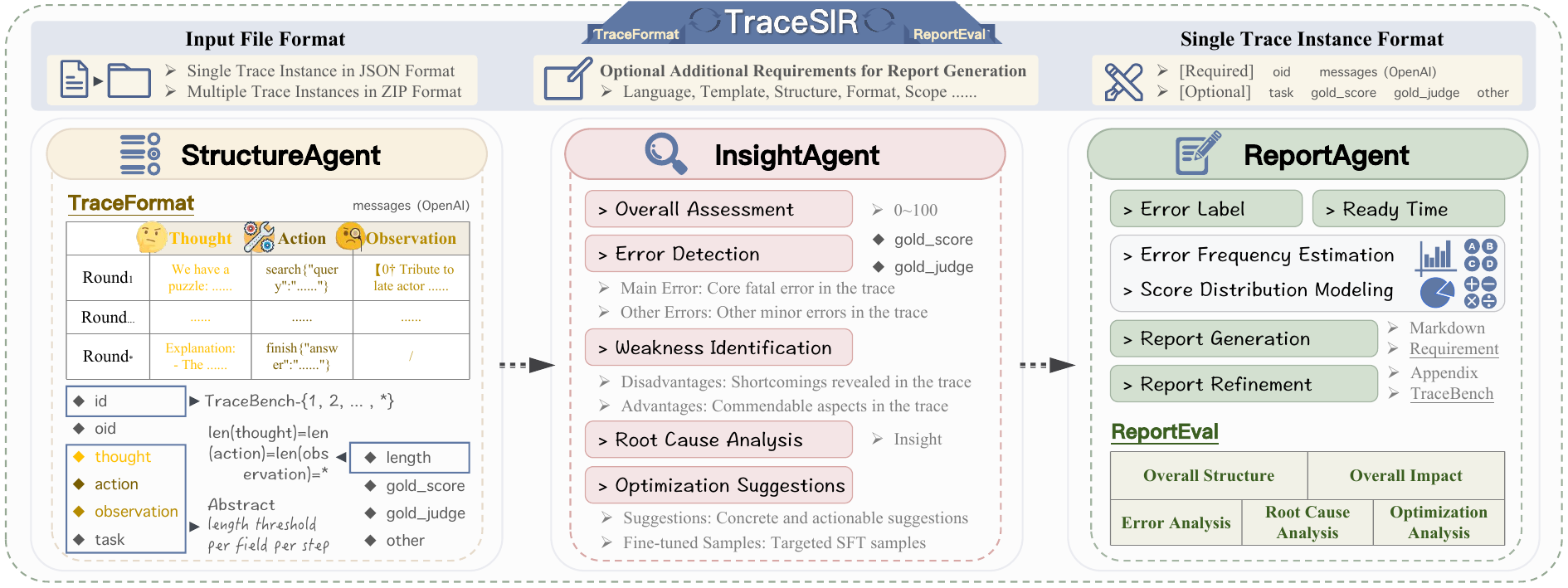}
  \caption{The overall architecture of \textbf{TraceSIR}.}
  \label{fig: framework}
  \vspace{-3mm}
\end{figure*}

\section{Related Work}




Most existing approaches \cite{browsecomp-deepresearch,tau2-functioncalling} for evaluating agentic systems remain outcome-oriented, which limits their ability to support issue localization and root cause analysis that depend on fine-grained behavioral context. Although some recent work attempts to incorporate partial intermediate signals, such as selected states \citep{dream,reward} or summarized steps \citep{trace,trail}, these abstractions inevitably discard substantial contextual information embedded in long execution traces and fail to support in-depth, insight-driven root cause analysis, making systematic diagnosis of agent behavior difficult. 

In practice, understanding agentic systems often requires analyzing multiple task instances and producing consolidated analysis reports, yet such settings have received limited attention in prior work. Existing approaches typically evaluate cases in isolation and aggregate results only at a superficial level \citep{trace,trail,dream}, which obscures recurring failure patterns across execution traces. Moreover, techniques based on naive summarization \citep{summary1,summary2,summary3} or clustering-style aggregation \citep{cluster1,cluster2,cluster3} struggle to preserve trace-level evidence, often resulting in overly coarse or weakly grounded reports.

In contrast, TraceSIR is designed specifically for structured analysis and reporting of agent execution traces. By abstracting long traces into structured representations and coordinating specialized agents for diagnosis and aggregation, TraceSIR enables scalable issue localization, root cause analysis, and actionable reporting across multiple cases.

\section{Method}

We propose \textbf{TraceSIR}, a multi-agent framework for structured analysis and reporting of agentic execution traces. As illustrated in Figure \ref{fig: framework}, TraceSIR coordinates three specialized components, i.e., {StructureAgent}, {InsightAgent}, and {ReportAgent}. Each component operates autonomously with tool invocation over structured traces, while the overall framework is designed to support both fine-grained per-case analysis and scalable cross-case reporting.

\subsection{TraceFormat}

The input to TraceSIR can be either a single execution trace in JSON format or multiple trace cases in ZIP format. Each trace case is represented as an object with a required \texttt{oid} field for trace identification, a \texttt{messages} field containing the execution trace in the standard OpenAI format, and several optional auxiliary fields.

Formally, for each trace case, the core execution trace stored in the \texttt{messages} field is represented as a sequence of messages $\mathcal{M} = \{ m_1, m_2, \dots, m_K \}$. Each message $m_k$ consists of a role (e.g., \texttt{user}, \texttt{assistant}, \texttt{tool}) and the corresponding content. This representation is widely adopted by modern agentic systems and naturally captures iterative reasoning, tool invocation, and environment feedback.



In addition to the \texttt{messages} field and an arbitrary trace identifier \texttt{oid}, a trace case may optionally include the following fields:
\begin{itemize}[left=1em, topsep=0pt, partopsep=0pt, parsep=0pt, itemsep=0pt]
\item \texttt{task}: a textual description of the task;
\item \texttt{gold\_score}: an automated evaluation score provided by external evaluators;
\item \texttt{gold\_judge}: textual feedback or error messages from automated assessment;
\item \texttt{other}: auxiliary or miscellaneous metadata.
\end{itemize}

TraceSIR also supports an optional additional \texttt{requirement} specification that constrains or guides the report generation process, such as preferences over language. In the absence of explicit specifications, reports are generated in Chinese by default, considering practical deployment settings. All optional fields are treated as auxiliary signals and are not required for TraceSIR to function.


To enable scalable and reliable analysis, TraceSIR introduces \texttt{TraceFormat}, a structured abstraction of OpenAI messages that preserves essential behavioral information while providing a compact and analyzable representation. Given an execution trace $\mathcal{M}$, we define a deterministic parsing function
\begin{equation}
\Phi: \mathcal{M} \rightarrow \mathcal{T} = \{ (t_i, a_i, o_i) \}_{i=1}^{N}
\end{equation}
which transforms the original message sequence into a structured trace $\mathcal{T}$ consisting of $N$ interaction rounds. In each round, $t_i$ means the agent’s \textbf{Thought}, capturing its intermediate reasoning expressed in assistant messages; $a_i$ denotes the \textbf{Action} selected by the agent, such as a tool invocation or explicit operation; and $o_i$ represents the resulting \textbf{Observation} returned by the environment or tool.



\texttt{TraceFormat} strictly preserves the temporal order and causal alignment among reasoning, action, and observation, and can be rendered as a three-column table to support both human inspection and automated analysis. Each trace case is assigned a standardized identifier with a \texttt{TraceBench} prefix through a newly introduced \texttt{id} field, ensuring consistent indexing and tool-based cross-case comparison throughout the analysis and reporting process. 

\subsection{StructureAgent}

While \texttt{TraceFormat} standardizes execution traces, long-horizon agent behaviors often contain verbose reasoning, long code snippets, or extensive tool outputs that exceed practical context limits for downstream analysis. {StructureAgent} addresses this bottleneck by performing trace-level abstraction.


Formally, given a structured trace $\mathcal{T}$, StructureAgent applies a length-aware abstraction operator parameterized by a threshold $\theta$ to produce a compressed trace $\mathcal{T}' = \mathcal{A}_{\theta}(\mathcal{T})$, where $\mathcal{A}_{\theta}$ selectively abstracts overlong elements while preserving essential behavioral information through automatic tool invocation. The abstraction is performed independently at the level of individual steps and across different fields, i.e., Thought, Action, and Observation, ensuring that only redundant or excessively verbose content is compressed, without obscuring critical signals required for subsequent analysis.

When necessary, StructureAgent also generates a compact abstraction of the \texttt{task} field. The resulting structured trace $\mathcal{T}'$ substantially reduces redundancy while remaining faithful to the original execution semantics, thereby enabling reliable and scalable downstream diagnosis.

\subsection{InsightAgent}

Built on the structured traces produced by StructureAgent, InsightAgent performs fine-grained, instance-level diagnostic analysis. Rather than relying solely on final task outcomes, it reasons over the entire structured execution trace to uncover behavioral issues and performance limitations.


Formally, given a structured trace $\mathcal{T}'$ and the associated \texttt{task} field $q$, InsightAgent produces a set of structured diagnostic outputs, denoted as 
$\mathcal{D} = \{ s, E, W, R, O \}$, by autonomously invoking analysis tools over the structured trace.
Here, $s$ represents an overall task completion assessment ranging from 0 to 100. 
$E$ captures detected errors in the execution trace and distinguishes between the primary, task-critical error and other secondary errors. 
$W$ denotes identified weaknesses revealed by the trace, while also optionally noting notable strengths when relevant. 
$R$ corresponds to root cause analysis, providing in-depth and insight-driven explanations of why the observed failures or limitations occur, and serves as a central diagnostic signal in our framework. 
Finally, $O$ consists of optimization suggestions, including concrete textual recommendations and fine-tuned samples to address the identified issues.


When available, \texttt{gold\_score} and \texttt{gold\_judge} are incorporated as auxiliary reference signals to support task completion assessment and error identification and interpretation. In their absence, InsightAgent derives all diagnostic outputs solely from trace-level behavioral evidence. All diagnostic outputs for each trace case are returned to the user in a structured format, supporting fine-grained, per-case analysis and enabling reliable aggregation and comparison across multiple cases.

\begin{table*}[t]
\centering
\small
\setlength{\tabcolsep}{6pt}
\begin{tabularx}{\linewidth}{lllc
>{\centering\arraybackslash}X
>{\centering\arraybackslash}X
>{\centering\arraybackslash}X
>{\centering\arraybackslash}X
>{\centering\arraybackslash}Xc}
\toprule
\textbf{Scenario} & \textbf{Method} & \textbf{Backbone} &
\cellcolor{red!10}{\textbf{Overall Score}} &
\textbf{OS} &
\textbf{EA} &
\textbf{RCA} &
\textbf{OA} &
\textbf{OI} & \cellcolor{gray!15}{\textbf{Ranking}}\\
\midrule

\multirow{4}{*}{Deep Research}
& ClaudeCode & GLM-5       & \cellcolor{red!10}{55.0} & 5.5 & 4.5 & \underline{5.5} & \underline{6.0} & {6.0} & \cellcolor{gray!15}{4}\\
& ClaudeCode & Claude-4.6 & \cellcolor{red!10}\underline{66.0} & {6.5} & \underline{7.5} & \underline{5.5} & \textbf{7.0} & {6.5} & \cellcolor{gray!15}{2}\\
& TraceSIR   & GLM-5       & \cellcolor{red!10}{65.0} & \underline{7.5} & 6.5 & \underline{5.5} & \underline{6.0} & \underline{7.0} & \cellcolor{gray!15}{3}\\
& TraceSIR   & Claude-4.6 & \cellcolor{red!10}\textbf{81.0} & \textbf{8.5} & \textbf{8.5} & \textbf{8.0} & \textbf{7.0} & \textbf{8.5} & \cellcolor{gray!15}{1}\\

\midrule
\multirow{4}{*}{Function Calling}
& ClaudeCode & GLM-5       & \cellcolor{red!10}{40.0} & 5.0 & 3.5 & {4.5} & {3.0} & {4.0} & \cellcolor{gray!15}{4}\\
& ClaudeCode & Claude-4.6 & \cellcolor{red!10}\underline{74.0} & \textbf{7.5} & \textbf{7.5} & \textbf{8.5} & \underline{6.5} & \underline{7.0} & \cellcolor{gray!15}{2}\\
& TraceSIR   & GLM-5       & \cellcolor{red!10}{53.0} & \underline{6.0} & \underline{6.0} & \underline{5.0} & {5.0} & {4.5} & \cellcolor{gray!15}{3}\\
& TraceSIR   & Claude-4.6 & \cellcolor{red!10}\textbf{77.0} & \textbf{7.5} & \textbf{7.5} & \textbf{8.5} & \textbf{7.5} & \textbf{7.5} & \cellcolor{gray!15}{1}\\

\midrule
\multirow{4}{*}{Agentic Coding}
& ClaudeCode & GLM-5       & \cellcolor{red!10}{57.0} & 6.5 & 5.5 & 5.5 & 6.0 & 5.0 & \cellcolor{gray!15}{4}\\
& ClaudeCode & Claude-4.6 & \cellcolor{red!10}\underline{77.0} & \underline{7.5} & \underline{8.0} & \underline{7.0} & \underline{7.5} & \underline{8.5} & \cellcolor{gray!15}{2}\\
& TraceSIR   & GLM-5       & \cellcolor{red!10}{62.0} & 5.5 & {6.5} & {6.5} & {6.5} & {6.0} & \cellcolor{gray!15}{3}\\
& TraceSIR   & Claude-4.6 & \cellcolor{red!10}\textbf{89.0} & \textbf{9.0} & \textbf{9.0} & \textbf{9.0} & \textbf{8.5} & \textbf{9.0} & \cellcolor{gray!15}{1}\\

\bottomrule
\end{tabularx}

\caption{Comparison of analysis report quality under the \texttt{ReportEval} protocol using \colorbox{red!10}{human evaluation}. TraceSIR is our system while ClaudeCode is the baseline.}
\label{tab: main_results_human}
\vspace{-3mm}
\end{table*}

\subsection{ReportAgent}

While InsightAgent focuses on individual task instances, practical diagnosis of agentic systems often requires coordinated analysis across multiple cases. {ReportAgent} operates at this higher level.

Formally, given a collection of diagnostic outputs $\{ \mathcal{D}_1, \dots, \mathcal{D}_M \}$ produced by InsightAgent for $M$ trace cases, ReportAgent autonomously invokes analysis tools to determine whether report generation is warranted based on the number of available cases and predefined triggering criteria. When activated, it performs targeted statistical analysis over the aggregated diagnostics, including error frequency estimation and score distribution modeling, to uncover recurring patterns and systematic issues.




For error analysis, ReportAgent summarizes the detected errors $E_i$ of each case into a canonical error label $\ell_i$, and estimates the frequency of each error type across $M$ cases as $P(\ell) = \frac{1}{M} \sum_{i=1}^{M} \mathbb{I}(\ell_i = \ell)$, where $\ell_i$ denotes the error label derived from $E_i$ for case $i$.  
For performance analysis, ReportAgent models the distribution of task completion scores over a set of predefined, disjoint score intervals $\mathcal{B} = \{ b_1, \dots, b_L \}$. The score distribution is estimated as $P(b) = \frac{1}{M} \sum_{i=1}^{M} \mathbb{I}(\hat{s}_i \in b)$, where $\hat{s}_i$ denotes the score used for aggregation, taking \texttt{gold\_score} when available and otherwise falling back to the score $s_i$ produced by InsightAgent.


ReportAgent integrates quantitative statistics with qualitative insights to generate comprehensive analysis reports in Markdown format by autonomously invoking reporting and analysis tools. When user-specified \texttt{requirement} fields are provided, they are incorporated during report generation; otherwise, reports are produced according to a default analysis schema.
After report generation, {ReportAgent} further refines the report by automatically identifying referenced trace cases through standardized \texttt{id} fields with the \texttt{TraceBench} prefix using a matching tool, and appending the corresponding structured trace data to the appendix. This process ensures transparency and traceability of the reported analyses.
By aggregating trace-level evidence across cases, TraceSIR produces coherent, evidence-grounded reports that support practical research and engineering decision-making.

\section{Experiments}

\begin{table*}[t]
\centering
\small
\setlength{\tabcolsep}{6pt}
\begin{tabularx}{\linewidth}{lllc
>{\centering\arraybackslash}X
>{\centering\arraybackslash}X
>{\centering\arraybackslash}X
>{\centering\arraybackslash}X
>{\centering\arraybackslash}Xc}
\toprule
\textbf{Scenario} & \textbf{Method} & \textbf{Backbone} &
\cellcolor{blue!10}{\textbf{Overall Score}} &
\textbf{OS} &
\textbf{EA} &
\textbf{RCA} &
\textbf{OA} &
\textbf{OI} & \cellcolor{gray!15}{\textbf{Ranking}}\\
\midrule

\multirow{4}{*}{Deep Research}
& ClaudeCode & GLM-5       & \cellcolor{blue!10}{82.7} & 8.7 & 8.0 & 7.3 & \textbf{9.0} & \underline{8.3} & \cellcolor{gray!15}{4}\\
& ClaudeCode & Claude-4.6 & \cellcolor{blue!10}\underline{90.0} & \textbf{9.3} & \underline{8.7} & \underline{9.0} & \textbf{9.0} & \textbf{9.0} & \cellcolor{gray!15}{2}\\
& TraceSIR   & GLM-5       & \cellcolor{blue!10}{88.0} & \underline{9.0} & 8.0 & \underline{9.0} & \textbf{9.0} & \textbf{9.0} & \cellcolor{gray!15}{3}\\
& TraceSIR   & Claude-4.6 & \cellcolor{blue!10}\textbf{91.3} & \textbf{9.3} & \textbf{9.0} & \textbf{9.3} & \textbf{9.0} & \textbf{9.0} & \cellcolor{gray!15}{1}\\

\midrule
\multirow{4}{*}{Function Calling}
& ClaudeCode & GLM-5       & \cellcolor{blue!10}{80.7} & 8.3 & 7.3 & \underline{7.7} & \underline{9.0} & \underline{8.0} & \cellcolor{gray!15}{4}\\
& ClaudeCode & Claude-4.6 & \cellcolor{blue!10}{88.0} & \underline{8.7} & 8.3 & \textbf{9.0} & \underline{9.0} & \textbf{9.0} & \cellcolor{gray!15}{3}\\
& TraceSIR   & GLM-5       & \cellcolor{blue!10}\underline{89.3} & \textbf{9.0} & \underline{8.7} & \textbf{9.0} & \underline{9.0} & \textbf{9.0} & \cellcolor{gray!15}{2}\\
& TraceSIR   & Claude-4.6 & \cellcolor{blue!10}\textbf{91.3} & \textbf{9.0} & \textbf{9.0} & \textbf{9.0} & \textbf{9.7} & \textbf{9.0} & \cellcolor{gray!15}{1}\\

\midrule
\multirow{4}{*}{Agentic Coding}
& ClaudeCode & GLM-5       & \cellcolor{blue!10}{58.7} & 6.0 & 4.0 & 5.7 & 8.0 & 5.7 & \cellcolor{gray!15}{4}\\
& ClaudeCode & Claude-4.6 & \cellcolor{blue!10}\underline{90.0} & \underline{9.0} & \textbf{9.0} & \textbf{9.0} & \textbf{9.0} & \textbf{9.0} & \cellcolor{gray!15}{2}\\
& TraceSIR   & GLM-5       & \cellcolor{blue!10}{84.7} & 8.3 & \underline{7.7} & \underline{8.7} & \textbf{9.0} & \underline{8.7} & \cellcolor{gray!15}{3}\\
& TraceSIR   & Claude-4.6 & \cellcolor{blue!10}\textbf{90.7} & \textbf{9.3} & \textbf{9.0} & \textbf{9.0} & \textbf{9.0} & \textbf{9.0} & \cellcolor{gray!15}{1}\\

\bottomrule
\end{tabularx}

\caption{Comparison of analysis report quality under the \texttt{ReportEval} protocol using \colorbox{blue!10}{LLM-as-a-judge}. TraceSIR is our system while ClaudeCode is the baseline.}
\label{tab: main_results_llm}
\vspace{-3mm}
\end{table*}

\subsection{TraceBench}





To evaluate the effectiveness of TraceSIR, we construct \textbf{TraceBench}, a unified benchmark of agentic execution traces collected from three representative real-world agentic benchmarks: BrowseComp (Deep Research) \citep{browsecomp-deepresearch}, Tau2Bench (Function Calling) \citep{tau2-functioncalling}, and SWE-bench (Agentic Coding) \citep{swebench-coding}. 
TraceBench includes 150 failed task instances of GLM-4.6 \footnote{https://huggingface.co/zai-org/GLM-4.6}, distributed across three scenarios.
Construction details are provided in Appendix \ref{sec: tracebench}.


\subsection{ReportEval}

To evaluate the quality and practical usefulness of analysis reports generated by TraceSIR, we propose \texttt{ReportEval}, a report-centric evaluation protocol for agentic execution trace analysis. \texttt{ReportEval} assesses each report along five dimensions with equal weight, producing both dimension-level scores and an overall quality score.

The five evaluation dimensions are defined as follows.
\textbf{Overall Structure} (OS) evaluates whether the report is well organized, coherent, and clearly grounded in the analyzed execution traces.
\textbf{Error Analysis} (EA) measures the correctness and trace support of the identified agent errors.
\textbf{Root Cause Analysis} (RCA) assesses whether the report provides insightful and well-justified explanations of the underlying causes of the observed errors.
\textbf{Optimization Analysis} (OA) evaluates the relevance, feasibility, and actionability of the proposed optimization suggestions.
\textbf{Overall Impact} (OI) provides a holistic assessment of the report’s usefulness from a practitioner’s perspective, reflecting its value for understanding agent behavior and supporting decision making. Each dimension is scored on a scale ranging from 0 to 10. The overall report score is computed as the equally weighted sum of the five evaluation dimension scores and normalized to a range from 0 to 100.

\subsection{Settings}


TraceSIR is instantiated with two backbone LLMs, GLM-5 \citep{glm5} 
and Claude-4.6 \footnote{https://www.anthropic.com/news/claude-opus-4-6}, and we adopt ClaudeCode \footnote{https://github.com/anthropics/claude-code} as the baseline model. 
For report evaluation, we adopt a hybrid setting that combines human assessment (6 expert LLM agent engineers/researchers) and automated LLM-based judging. Human evaluators were blinded to the underlying methods and model identities. Further experimental settings and evaluation details are provided in Appendix \ref{sec: setting}.


\subsection{Results}

From Table \ref{tab: main_results_human} and Table \ref{tab: main_results_llm}, it can be observed that TraceSIR consistently and substantially outperforms ClaudeCode across all three agentic scenarios, demonstrating clear advantages in overall report quality and fine-grained diagnostic capability.
From the perspective of human evaluation, TraceSIR yields consistent and sizable improvements over ClaudeCode with both backbone models. When using the weaker GLM-5 backbone, TraceSIR improves the overall score by 10.0\%, 13.0\%, and 5.0\% in Deep Research, Function Calling, and Agentic Coding, respectively. With the stronger Claude-4.6 backbone, TraceSIR further improves performance by 15.0\%, 3.0\%, and 12.0\% across these scenarios. Averaged across scenarios and backbones, TraceSIR achieves a 9.7\% relative improvement under human evaluation, indicating a clear expert preference for its reports.
Averaged across the five 10-point evaluation dimensions, TraceSIR further achieves mean improvements of 0.9, 1.3, 1.2, 0.8, and 0.9 points, respectively, with particularly pronounced gains in error analysis and root cause analysis, demonstrating its strengthened diagnostic capability and its effectiveness in supporting systematic analysis and report generation.



Results from LLM-as-a-judge evaluation exhibit a highly consistent trend. With the GLM-5 backbone, TraceSIR improves the overall score by 5.3\%, 8.6\%, and 26.0\%, respectively. When paired with Claude-4.6, the corresponding improvements are 1.3\%, 3.3\%, and 0.7\%. On average, TraceSIR yields a 7.5\% relative improvement over ClaudeCode under LLM-based evaluation. Notably, the largest gains occur in more challenging settings, particularly Agentic Coding with weaker backbone models, highlighting TraceSIR’s robustness in producing diagnostically meaningful reports under constrained model capacity.
Although the absolute scores differ between the two evaluation settings, the overall conclusions remain consistent. Human experts tend to provide more conservative and stringent assessments, whereas the LLM judge assigns relatively higher scores, reflecting a more permissive evaluation behavior. Despite this difference in score calibration, both evaluation methods produce highly aligned rankings and converge on the same qualitative conclusion that TraceSIR generates more structured, comprehensive, and actionable reports than ClaudeCode across all scenarios, with consistent advantages observed across all dimensions defined in the \texttt{ReportEval} protocol.



\section{Conclusion}

We presented \textbf{TraceSIR}, a framework for structured analysis and reporting of agentic execution traces. By introducing \texttt{TraceFormat} to abstract long execution traces while preserving critical behavioral information, TraceSIR enables scalable issue localization and insight-driven root cause analysis. The coordinated design of StructureAgent, InsightAgent, and ReportAgent further supports both instance-level diagnosis and cross-case aggregation, producing coherent and actionable analysis reports.
Experiments on \textbf{TraceBench} show that TraceSIR generates high-quality reports aligned with practical research and engineering needs. 

\section*{Limitations}

Despite its effectiveness, TraceSIR has several limitations. First, TraceSIR relies on LLM-based agents for trace abstraction, diagnostic reasoning, and report generation, and the quality of the resulting analyses may be influenced by the capabilities of the underlying language models, particularly in challenging domains such as complex coding tasks or highly specialized technical settings.
Second, while TraceSIR is designed to support scalable cross-case analysis, its current report generation strategy assumes a moderate number of task instances. Applying TraceSIR to very large collections of traces may require additional mechanisms for hierarchical aggregation or incremental reporting to maintain report clarity and efficiency.
Finally, TraceSIR may incur relatively long response times and high token consumption, and its reports may vary due to the inherent randomness of LLMs. These factors limit its efficiency and reproducibility in practical use, and mitigating latency, cost, and variability remains an important direction for future work.

\section*{Ethics and Broader Impact}

This work is conducted in accordance with the ACM Code of Ethics. TraceSIR is a system for structured analysis and reporting of agentic execution traces, designed to help researchers and engineers understand agent behaviors, diagnose failures, and identify root causes in complex, long-horizon agentic systems. By transforming raw execution traces into structured, analyzable representations and aggregating evidence across multiple task instances, TraceSIR aims to improve transparency, reliability, and usability of agentic systems.

\paragraph{Data and Privacy.}
TraceSIR operates on execution traces generated by agentic systems and does not require access to personal data by design. The datasets used in our experiments are publicly available benchmarks, and all traces are collected from synthetic or benchmark-defined tasks. Nevertheless, when TraceSIR is applied in real-world deployments, execution traces may contain sensitive information such as proprietary prompts, tool outputs, or user-provided content. In such settings, appropriate data handling practices, including access control, anonymization, and secure storage, should be followed.

\paragraph{Potential Benefits.}
TraceSIR is intended to support researchers and engineers who develop, deploy, and maintain agentic systems, by enabling systematic issue localization, root cause analysis, and cross-case reporting that are difficult to achieve through manual inspection. By providing coherent and evidence-grounded analysis reports with actionable optimization suggestions, TraceSIR may improve debugging efficiency, support informed engineering decisions, and contribute to safer, more reliable, and more interpretable agentic systems.

\paragraph{Potential Risks and Misuse.}
TraceSIR relies on LLM-based agents for trace abstraction, diagnostic reasoning, and report generation. As a result, its outputs may reflect limitations, biases, or inaccuracies of the underlying models. Over-reliance on automatically generated analyses without human verification could lead to incorrect conclusions or suboptimal system modifications. TraceSIR is therefore designed as a decision-support tool rather than an autonomous authority, and its outputs are intended to be interpreted by practitioners.

\paragraph{Broader Impact.}
More broadly, this work aims to promote principled and transparent analysis practices for agentic systems. While TraceSIR can be used to improve system quality and robustness across a range of research and engineering settings, it may also be applied to analyze large-scale agent deployments. Responsible use of such systems requires careful consideration of transparency, accountability, and data governance, particularly in real-world applications. We encourage future work to further explore safeguards, evaluation practices, and governance mechanisms for responsible use of automated trace analysis and reporting systems.

\bibliography{custom}

\clearpage

\begin{figure}[t]
  \centering
  \includegraphics[width=\linewidth]{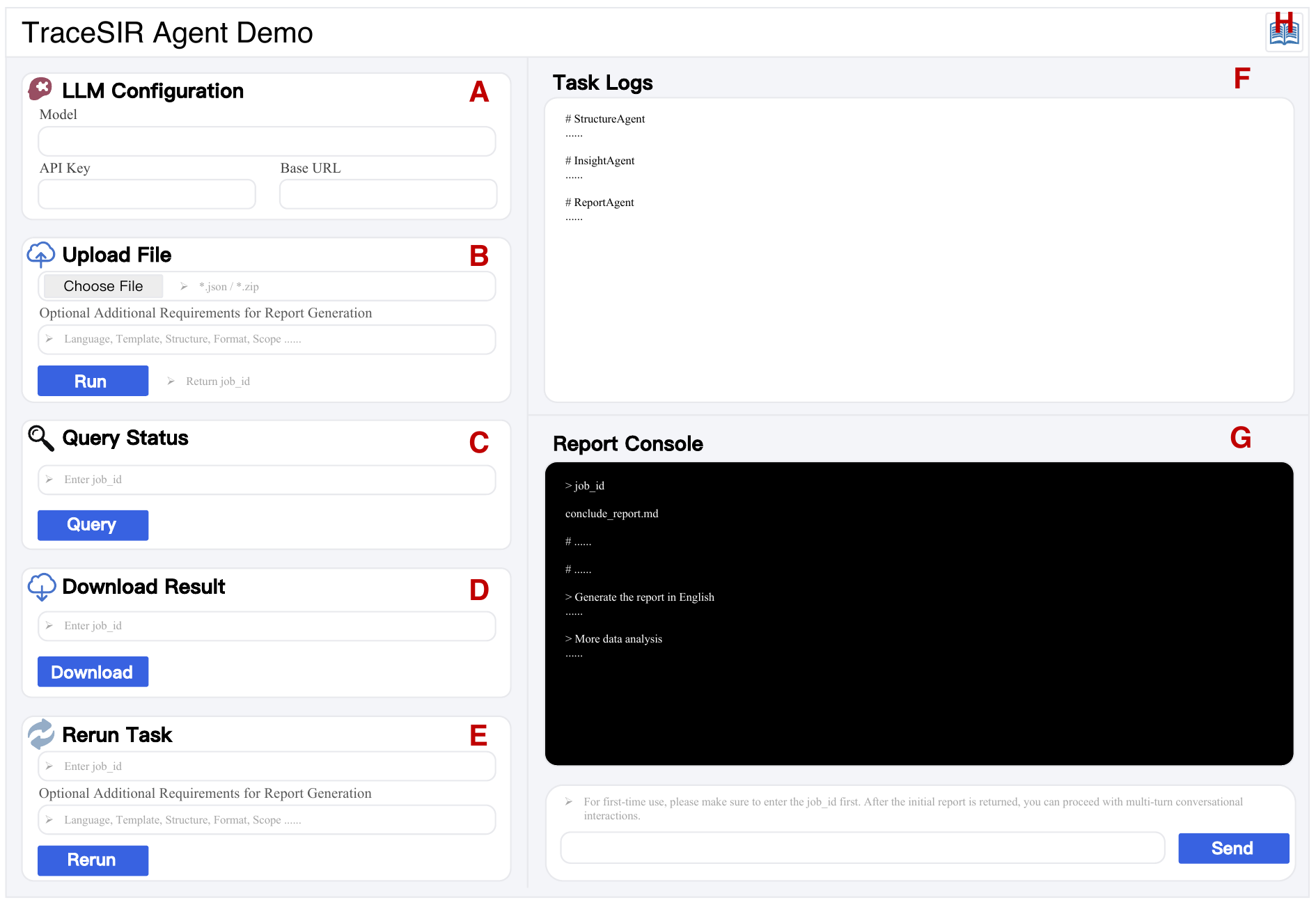}
  \caption{The demonstration of TraceSIR.}
  \label{fig: demo}
\end{figure}

\appendix

\section{TraceBench Construction}
\label{sec: tracebench}

For the BrowseComp, Tau2Bench, and SWE-bench benchmarks, we run GLM-4.6 \footnote{https://docs.bigmodel.cn/cn/guide/models/text/glm-4.6} as the target agent model on the official test split and retain the complete execution traces in the standard OpenAI message format. Task performance is evaluated using the official evaluation protocols provided by each benchmark, yielding a binary \texttt{gold\_score}.  
To focus on diagnostic analysis of agent failures, we randomly sample 50 failure cases from each benchmark, where the \texttt{gold\_score} equals 0. In total, TraceBench contains 150 failed task instances across the three scenarios. All collected execution traces are used as inputs to TraceSIR for structured analysis and reporting of agentic execution traces.

\begin{figure}[t]
  \centering
  \includegraphics[width=\linewidth]{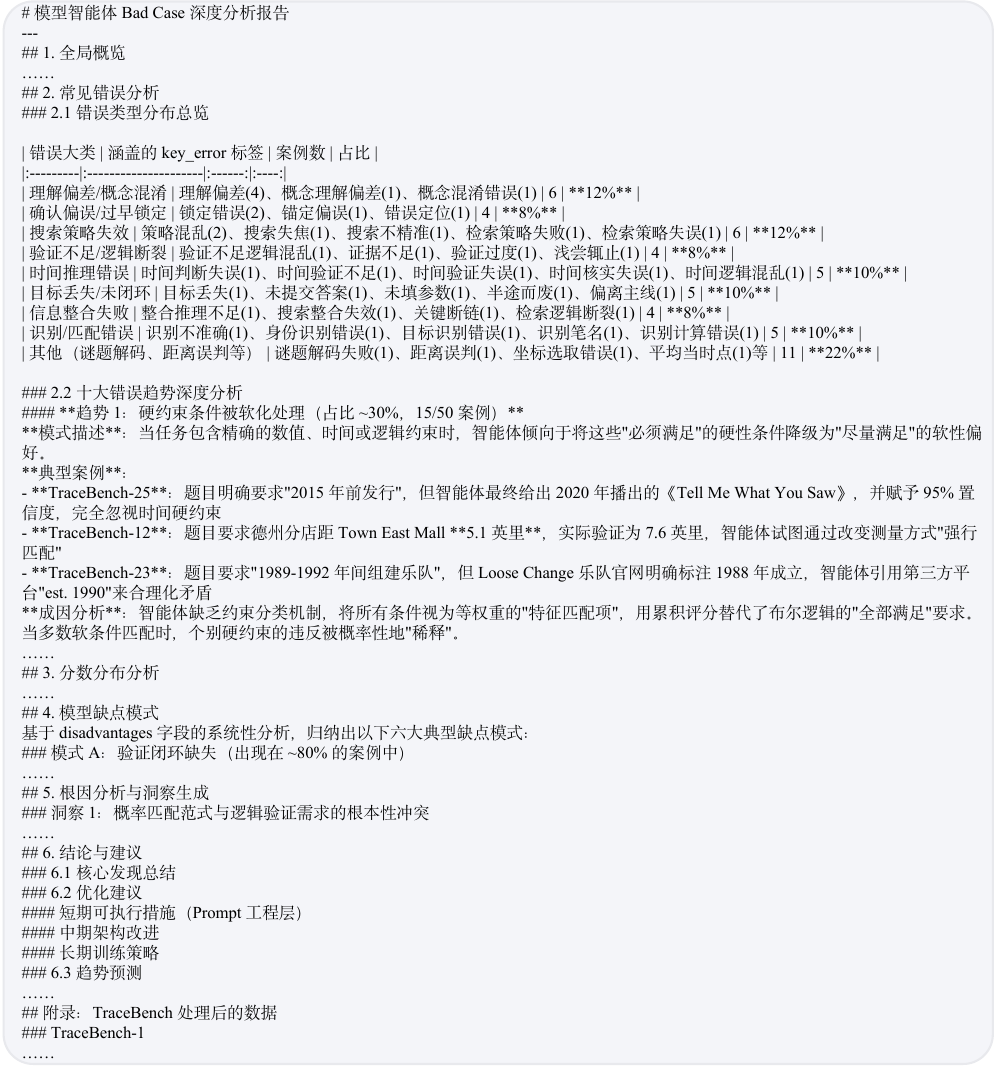}
  \caption{An excerpt from a Chinese analysis report generated by TraceSIR using Claude-4.6 on the Deep Research scenario of BrowseComp.}
  \label{fig: example}
\end{figure}

\section{Experimental Settings}
\label{sec: setting}

We evaluate TraceSIR under a controlled experimental setting. 
The length threshold $\theta$ for trace abstraction is set to 100 words or 1,000 characters.
During processing, report generation is automatically triggered when the number of processed cases reaches a multiple of 10. For evaluation consistency, however, we compare only the final reports generated after all 50 cases in each dataset have been processed. For each agentic scenario in TraceBench, all methods operate on the same set of 50 sampled failure cases with identical execution traces, producing one report per scenario for comparison.
Notably, in practical deployments and our demonstration system, in addition to the default report generation setting described above, we enforce the generation of a final analysis report regardless of the number of input cases, in order to improve user experience. As a result, a reference report is always produced at the end of analysis, even when only a single case is provided.

For report evaluation, we adopt a hybrid setting that combines expert human assessment with automated LLM-based judging. 
Specifically, six domain experts with extensive experience in agentic systems manually evaluate the generated reports following the \texttt{ReportEval} protocol.
The final human evaluation score is obtained by averaging the ratings across experts, with each scenario evaluated by at least two experts.
In addition, we adopt an LLM-as-a-judge approach under the \texttt{ReportEval} protocol using GPT-5 \footnote{https://developers.openai.com/api/docs/models/gpt-5}, running the evaluation three times and averaging the scores to improve stability, thereby enabling scalable and consistent assessment complementary to human evaluation.

\begin{table*}[t]
\centering
\small
\begin{tabularx}{\linewidth}{l X c c}
\toprule
\textbf{Category} & \textbf{Capability} & \textbf{TraceSIR} & \textbf{ClaudeCode} \\
\midrule

\multirow{5}{*}[-1.5ex]{System Design}
& 1. Multi-agent architecture explicitly designed for analysis and reporting, with separated roles for trace structuring, diagnosis, and reporting & $\checkmark$ & $\triangle$ \\
& 2. Fine-grained, per-case diagnostic outputs with structured fields & $\checkmark$ & $\triangle$ \\
& 3. Native cross-case aggregation and statistical analysis over all traces & $\checkmark$ & $\triangle$ \\
& 4. Incremental analysis with resume support for large trace collections & $\checkmark$ & $\times$ \\
& 5. Configurable analysis and report constraints via explicit settings & $\checkmark$ & $\triangle$ \\

\midrule
\multirow{5}{*}{Methodology}
& 1. Full execution trace modeling with explicit Thought--Action--Observation & $\checkmark$ & $\times$ \\
& 2. Structured trace abstraction preserving causal and temporal dependencies & $\checkmark$ & $\times$ \\
& 3. Insight-driven root cause analysis beyond surface-level error description & $\checkmark$ & $\triangle$ \\
& 4. Cross-step causal reasoning over long-horizon agent behaviors & $\checkmark$ & $\times$ \\
& 5. Failure-centric analysis designed for systematic diagnosis & $\checkmark$ & $\triangle$ \\

\midrule
\multirow{5}{*}{Reporting}
& 1. Multi-dimensional analysis reports with explicit diagnostic categories & $\checkmark$ & $\triangle$ \\
& 2. Trace-level evidence linking conclusions to specific execution steps & $\checkmark$ & $\times$ \\
& 3. Actionable optimization suggestions grounded in diagnosed root causes & $\checkmark$ & $\triangle$ \\
& 4. Generation of targeted SFT samples for downstream model improvement & $\checkmark$ & $\times$ \\
& 5. Robust analysis report quality under weaker backbone language models & $\checkmark$ & $\times$ \\

\bottomrule
\end{tabularx}
\caption{Capability-level comparison between TraceSIR and ClaudeCode. 
$\checkmark$ indicates native and systematic support, 
$\triangle$ indicates partial or non-systematic support, 
and $\times$ indicates lack of support.}
\label{tab: trace_vs_claudecode}
\end{table*}

\section{TraceSIR Demonstration}

Figure~\ref{fig: demo} presents an end-to-end demonstration of TraceSIR, illustrating how the system supports structured analysis and reporting of agentic execution traces in practice. The demo provides a unified interface for configuring language models, submitting execution traces, monitoring analysis progress, retrieving results, and interactively refining generated reports. It is designed to reflect realistic usage scenarios of TraceSIR and to support transparent and reproducible analysis workflows.

The demo interface consists of the following components, labeled from A to H in Figure~\ref{fig: demo}.

\paragraph{A. LLM Configuration.}
This component allows users to configure the underlying model, including model name, API key, and base URL, enabling flexible deployment across different LLM backends.

\paragraph{B. Upload Analysis File.}
Users can submit execution traces in JSON format for a single case or in ZIP format for multiple cases. This component also supports an optional additional user-specified \texttt{requirement} field that constrains or guides report generation. Upon submission, a unique job identifier is returned for subsequent operations.

\paragraph{C. Query Task Status.}
Given a job identifier, this component retrieves the current execution status and metadata of the analysis task, including processing progress, timestamps, and detailed logs.

\paragraph{D. Download Analysis Results.}
This component enables downloading the analysis outputs. The returned archive contains all processed trace files and generated analysis reports.

\paragraph{E. Rerun Task.}
This component supports re-executing an existing task using the same input traces. It allows users to resume interrupted analyses, which is particularly useful when processing large collections of trace cases. Previously analyzed cases are not reprocessed, and the system continues only with unfinished cases. After all cases have been analyzed, users may also rerun the task to regenerate analysis reports without reprocessing completed traces, optionally updating report generation requirements.

\paragraph{F. Task Logs.}
This panel displays execution logs produced during trace processing and analysis, providing transparency into the behavior of {StructureAgent}, {InsightAgent}, and {ReportAgent}.

\paragraph{G. Report Console.}
After report generation, users can interactively refine the analysis report through multi-turn dialogue, enabling targeted revisions without rerunning the full analysis process.

\paragraph{H. Documentation.}
This entry provides access to system documentation, including supported functionalities, input formats, and operational guidelines for using TraceSIR.

\section{Example Analysis Report}

To illustrate the qualitative characteristics of reports generated by TraceSIR, we present a representative excerpt from an analysis report, as shown in Figure~\ref{fig: example}. 
The report is generated in Chinese, reflecting practical deployment settings and user requirements.
The complete reports, along with all related data, are available in our repository.

\section{TraceSIR VS ClaudeCode}

To better contextualize the design choices and empirical results of TraceSIR, we provide a capability-level comparison with ClaudeCode, a strong baseline for agentic analysis. Table~\ref{tab: trace_vs_claudecode} summarizes the key differences between the two systems across system design, methodology, and reporting capabilities, highlighting TraceSIR’s strengths in trace-centric diagnosis and cross-case reporting.

\end{document}